# Time-series Imputation and Prediction with Bi-Directional Generative Adversarial Networks

Mehak Gupta,[1] Rahmatollah Beheshti[1,2]

[1]Computer and Information Sciences, University of Delaware, Newark, DE, USA
[2]Epidemiology Program, University of Delaware, Newark, DE, USA
{mehakg, rbi}@udel.edu

## Abstract

Multivariate time-series data are used in many classification and regression predictive tasks, and recurrent models have been widely used for such tasks. Most common recurrent models assume that time-series data elements are of equal length and the ordered observations are recorded at regular intervals. However, real-world time-series data have neither a similar length nor a same number of observations. They also have missing entries, which hinders the performance of predictive tasks. In this paper, we approach these issues by presenting a model for the combined task of imputing and predicting values for the irregularly observed and varying length time-series data with missing entries. Our proposed model (Bi-GAN) uses a bidirectional recurrent network in a generative adversarial setting. The generator is a bidirectional recurrent network that receives actual incomplete data and imputes the missing values. The discriminator attempts to discriminate between the actual and the imputed values in the output of the generator. Using the available data in its entirety, our model learns how to impute missing elements in-between (imputation) or outside of the input time steps (prediction), hence working as an effective any-time prediction tool for time-series data. Our method has three advantages to the state-of-the-art methods in the field: (a) single model can be used for both imputation and prediction tasks; (b) it can perform prediction task for time-series of varying length with missing data; (c) it does not require to know the observation and prediction time window during training which provides a flexible length of prediction window for both long-term and short-term predictions. We evaluate our model on two public datasets and on another large real-world electronic health records dataset to impute and predict body mass index (BMI) values in children and show its superior performance in both settings.

## Introduction

Multivariate time-series prediction is widely used in many applications like economics, meteorology, communications, and control engineering. It is also studied heavily in biomedical fields for various applications such as estimating future health outcomes, monitoring health trajectories, and treatment responses. Among other issues, there are several that make working with real-world time-series data challenging, including missing entries and unequal lengths of time-series. Real-world time-series are also often not collected at regular intervals. Time intervals might be different within a sample or in between samples. These issues can hinder the performance of any time-series classification or regression tasks.

We approach these major issues of using time-series data for prediction, where data is of varying lengths, with irregularly observed and has missing entries. We propose a model that can perform both imputation and prediction in time-series data simultaneously. Our proposed model, named Bi-GAN, uses a bidirectional recurrent neural network (RNN) in a generative adversarial network (GAN) setting to learn the distribution of the available data. The motivation for using a GAN architecture in this setting is the flexibility in defining desired GAN loss functions that can effectively guide the model to learn the overall distribution of the time-series data. Using our proposed method, the model can learn from the available dataset in its entirety, where different time-series have missing entries at different time-points and have different lengths. We align all the time-series in the data and pad shorter length time-series to have the same length as longer time-series. We treat padded values as missing values that are filled during the imputation task. Our model learns from all the observed entries to impute any in-between missing entries, and also to fill-in future entries for shorter time-series by learning from longer time-series. Therefore, it can be used as an effective "any-time prediction tool." This setting does not require to define the observation and prediction window at the time of training. This solves another issue of time-series prediction, where observation and prediction windows need to be pre-defined. As we discuss in the next section, there is a large body of literature dedicated to addressing missingness and irregular patterns in time-series datasets. In this work, we show that our proposed model obtains superior performance for imputation and prediction tasks. In particular, we make the following technical contributions:
- We present a bidirectional RNN model in a generative adversarial setting for solving both imputation and prediction tasks in time-series data.
- We evaluate our model on two common time-series datasets (UCI's air-quality and MIMIC-III), as well as on another large electronic health record (EHR) dataset related to longitudinal records from around 70,000 patients in 10 years. Experimental results show that our model outperforms the state-of-the-art models for both imputation and prediction tasks.

## Related Work

Real-world datasets are prone to missingness due to reasons such as a fault in data collecting device, human error, and sample destruction. Missing data reduces the utility of large datasets in predictive tasks. There has been a substantial amount of research in developing methods that can handle missing values. Case deletion methods, where instances with missing elements are deleted, are among the simplest methods that may ignore some important information (Kaiser 2014; Silva and Zárate 2014). Also, interpolation methods (Fung 2006; Kreindler and Lumsden 2012; Silva and Zárate 2014) that use local interpolations to impute missing values can discard important temporal patterns across time. Multivariate Imputation by Chained Equations (MICE) (Azur et al. 2011) is perhaps the most popular method in this category, which uses a chained equation over various iterations to estimate missing values after an arbitrary initialization. Autoregressive methods (Harvey 1990), like ARIMA and SARIMA, fit a parameterized stationary model. Machine learning models like KNN (Tak et al. 2016), expectation maximization (Batista and Monard 2003), matrix factorization (Hastie et al. 2015; Mazumder et al. 2010) multi-layer perceptron (MLP) (Chen et al. 2001), and random forest (Li et al. 2018) are also among the common methods. However, almost all these methods do not consider the temporal dependencies between variables.

Because of RNNs' inherent capabilities in recognizing sequential patterns in time-series, many RNN-based methods have been presented for imputing time-series data. To name a few, Li et. al. (2018) use a long-short term memory with a support vector regressor to impute missing values in real-time traffic monitoring data. Mulyadi et al. (2020) consider the correlations between the input features during the training and perform time series classification along with imputation, and Cao et al. (2018) use a bi-directional recurrent dynamics architecture to impute data in electronic health record data. Many of such RNN-based methods have been used for time-series prediction. For instance, Choi et al. (2016) use RNNs for heart failure prediction, Jiang et al. (2018) use RNNs for short-term urban mobility prediction, and Chen et al. (2019) and Yao et al. (2019) use RNNs for traffic prediction.

Most existing RNN-based prediction models use same-length time-series and divide the time-series into a fixed observation and prediction window. These models are generally trained using the values in the observation window to predict values in the prediction window. For instance, Choi et al. (2016) use RNNs for heart failure prediction 6 months ahead from data observed in 3-12 months and Reddy et al. (2018) use RNNs for predicting patients' readmission for lupus patients within the next 30 days. Unlike this common prediction setting, the imputation task does not require to divide time-series into observation and prediction windows. In our work, we approach the problem of prediction in time-series as an imputation of future values. This technique helps us achieve flexible observation and prediction window, which can be used for both short-term and long-term predictions while working with time-series of varying length.

Recently generative adversarial networks have been also used for imputation algorithms. One of the popular applications of GANs has been using such architectures to generate synthetic datasets (Choi et al. 2017; Frid-Adar et al. 2018). GANs' ability to generate synthetic samples similar to the actual data can be utilized to impute missing values in the data, such that imputed values are close to the actually observed values. As an example study, Yoon et al. (Yoon et al. 2018) have used GANs with a hint vector to impute missing data in multivariate data. Several works have also used RNN architectures to implement the GAN components for imputing time-series data (Esteban et al. 2017). A good example of such studies is the work by Luo et al. (2018), where they use a modified gated recurrent unit in a GAN structure to impute missing data in multivariate time-series data. As another example, Yahi et. al. (2017) use GANs to generate synthetic time-series data. In both studies, the prediction task is performed using the imputed (synthetic) data and this method is used to evaluate the predictive power of the imputed data. In most of this type of studies, the imputation and prediction tasks are performed asynchronously by training two parts of the model separately. An example of this explicit separations is the two-stage framework including a missing data imputation and disease prediction proposed by Hwang et al (2017).

Building on this recent line of research, in this paper, we use a bi-directional RNN architecture in a generative adversarial setting to perform both imputation and prediction in time-series data. While not using a GAN architecture, a close study to our work is the BRITS-I (Cao, Wang, Li, Zhou, Li and Li 2018) model, which imputes missing data in time-series data using bi-directional RNNs. This model does not work well for the prediction task, as the imputed values are the simple mean of the forward and backward imputation from the bi-directional recurrent model. Our method does not have such an issue, as it uses weight parameters to take the weighted sum of the forward and backward imputed values. This improves the performance of both prediction and imputation tasks.

## Problem Setup

Suppose a multivariate time-series data $X$ in a d-dimensional space and observed over $n$ timestamps $T = (t_0, t_1, .... t_{n-1})$, is denoted by $X = \{x_0, x_1, ..., x_{n-1}\} \in \mathbb{R}^{n \times d}$, where $x_i$ is the $i$th observation vector of $X$, and $x_{ij}$ is the $j$th feature in the $x_i$ observation. We call $X$ the data vector, and also define the mask vector M such that it indicates which components of $X$ are missing in the following way:

$$m_{ij} = \begin{cases} 0 & \text{if } x_{ij} \text{ is missing} \\ 1 & \text{otherwise} \end{cases}$$

By following this formulation, we will have $M = \{m_0, m_1, ..., m_{n-1}\} \in \{0,1\}^{n \times d}$, where $m_i$ is the $i$th vector of M corresponding to the $i$th observation of $X$, and $m_{ij}$ is $j$th value of $m_i$ corresponding to the $j$th feature of $x_i$. Values in $X$ could be missing randomly, and at random times. To record the time gap between the values in X in the forward direction, we define $\delta_{ij}^f$ as the time gap vector such that:

$$\delta_{ij}^f = \begin{cases} t_i - t_{i-1}, & \text{if } m_{ij-1} == 1 \\ \delta_{ij-1}^f + t_i - t_{i-1}, & \text{if } m_{ij-1} == 0, i > 0 \\ 0, & i == 0 \end{cases}$$

This way, $\delta_{ij}^f$ will be the difference in the time of the last observed value and the current time step in the forward direction. Similarly, we calculate $\delta_i^b$ which is the time gap between the last observed value and the current time step in the backward direction. For calculating $\delta_i^b$, we reverse the timestamps to $T' = (t_{n-1}, .... t_1, t_0)$. Figure 1 shows the calculation of the time gaps $\delta^f$ and $\delta^b$ for the data vector $X$, where the time-series are recorded at $(t_1, t_2, t_3, t_4, t_5) = (1, 2, 3, 4, 5)$. Each column in $X$ represents the $x_i$th vector, and each row refers to the $j$th feature recorded at all time-series. Empty cells show the missing entries.

## Method

Following existing work on using RNNs and GAN-based architectures for analyzing time-series data, we present our model, Bi-GAN, which is an architecture based on a GAN and works internally using bidirectional recurrent dynamics. This model consists of the generator ($G$) and the discriminator ($D$). We use bidirectional LSTM cell layers to introduce recurrent components in $G$ and $D$. $G$ uses a bidirectional recurrent dynamical system, where each value is generated twice; once by its predecessor values in the forward direction and another time by its successor values in the backward direction.

For simplicity, we present our method for imputing or predicting the values of a univariate time-series variable, which we call a target variable. We still use multivariate time-series as input. The same method should be extendable to the case where the target is also multivariate. As an example, consider the problem of estimating an individual's body weight using all the health information (containing many features) of that individual in the past 10 years, versus estimating her body weight, height, and blood pressure using the same input. This way, our goal is to fill-in missing values for a single target variable ($x$) in $X$, and the output of $G$ would be a univariate time-series of values of that target variable. As shown in Figure 1, $G$ takes multivariate time-series data $X$ as input, and outputs the imputed values for the target variable $x$'s univariate time-series (shown by $\tilde{x}$). The generated values in $\tilde{x}$ are then replaced by the actual values (existed in the input) to obtain the output $\bar{x}$:

$$\bar{x} = x \odot m + \tilde{x} \odot (1 - m) \qquad (1)$$

where x and $\tilde{x}$ are the univariate time-series of the target variable in input $X$, and the generated output respectively, $m$ is the mask vector for the target variable, and $\odot$ shows the dot product operation.

$G$ consists of a recurrent component for time-series representation, and a regression component to generate the final output from the output of the recurrent layers. For $G$, we use one layer of bidirectional recurrent cells. Consider a standard RNN cell represented by,

$$h_i = \sigma(W_h h_{i-1} + U_h x_i + b_h) \qquad (2)$$

where $\sigma$ is the sigmoid function, $W_h, U_h$ and $b_h$ are model parameters and $h_{i-1}$ is the hidden state from the previous time steps. We use a bidirectional recurrent layer, to obtain the outputs – one from the forward direction $h_i^f$, and one from the backward direction $h_i^b$, using $\gamma_i^f$ and $\gamma_i^b$. These two parameters ($\gamma_i^f$ and $\gamma_i^b$) are the temporal decay factors calculated using $\delta_i^b$ and $\delta_i^f$ respectively and added to the hidden state calculations. We extend the standard recurrent component shown in Equation 2 by,

$$h_i^f = \sigma(W_h^f h_{i-1}^f \odot \gamma_i^f + U_h^f \overline{x_i} + b_h^f) \qquad (3)$$

$$\gamma_i^f = \exp(-\max(0, W_\gamma^f \delta_i^f + b_\gamma^f)) \qquad (4)$$

where $\overline{x_i}$ is the imputed values as shown in Eq. 1 at time $i$, and $W_\gamma^f$ and $b_\gamma^f$ are model parameters. Similar calculations to the forward case (Equation 3 and 4 above) are also used for the backward direction in the bidirectional recurrent layers (for calculating $h_i^b$ and $\gamma_i^b$). Here, $\gamma_i^f$ and $\gamma_i^b$ are calculated in such a way that $\gamma_i^f, \gamma_i^b \in (0,1]$, higher the $\delta_i^f$ and $\delta_i^b$ smaller the values of $\gamma_i^f$ and $\gamma_i^b$. By using the decay factors, we can calculate the confidence in forward and backward imputed values. For instance, in Figure 1 $x_{41}$ is missing and the observed successor value $x_{51}$ is closer than

https://github.com/mehak25/BiGAN

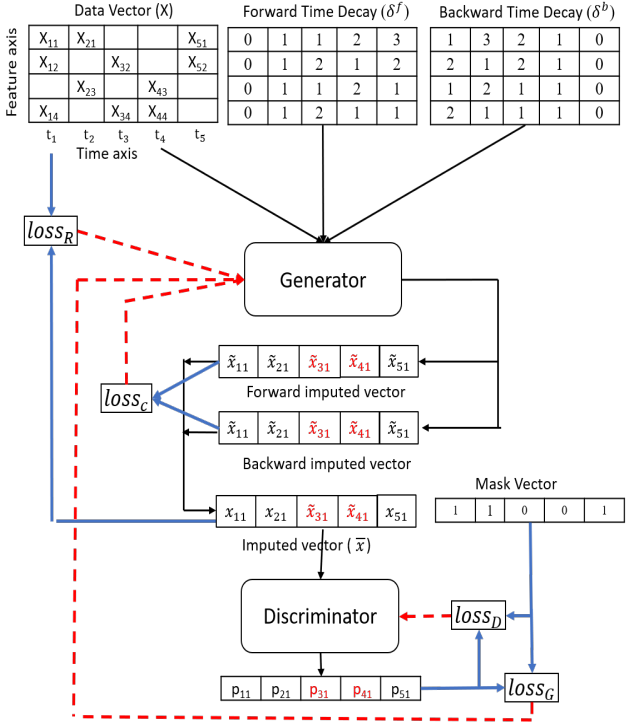

Figure 1: Bi-GAN architecture overview

the observed predecessor value $x_{21}$. In this case, the value generated for $x_{41}$ in the backward imputed vector should be more reliable.

The regression component of the generator is a fully-connected layer that generates the values for the target variable using the output of the recurrent layer. The regression component for forward direction is represented as,

$$\widetilde{x_1^f} = W_x^f h_{i-1}^f + b_x^f \quad (5)$$

where $\widetilde{x_i^f}$ is the generated value in forward and $W_x^f$ and $b_x^f$ are the model parameters. $\widetilde{x_i^b}$ will be calculated in backward direction in the same way as $x_i^f$ shown in Eq. 5

As shown in Figure 1, both the forward and backward imputed vectors are combined to calculate the final generated values:

$$\tilde{x} = \lambda_i^f x_i^f + \lambda_i^b \widetilde{x_i^b} \quad (6)$$

where $\lambda_i^f$ and $\lambda_i^b$ are two combination factors. These combination factors are trained as model parameters based on the time gap values in both forward and backward directions ($\delta_i^f$ and $\delta_i^b$). They help to control the influence of the forward and backward imputed values based on how far the last observed value in both directions was. The combination factor in the forward direction is calculated as,

$$\lambda_i^f = \exp(-\max(0, W_{\lambda f} \delta_i^f + b_{\lambda f})) \quad (7)$$

where $W_{\lambda f}, b_{\lambda f}$ are trained jointly with other parameters of the model. $\lambda_i^b$ is calculated for the backward direction in a similar way as $\lambda_i^f$ in Eq. 7. As we consider $\lambda_i^f, \lambda_i^b \in (0,1]$, higher the $\delta_i^f$ and $\delta_i^b$ smaller the values of $\lambda_i^f$ and $\lambda_i^b$.

The model is trained using four different losses. To ensure that the output generated by the generator $G$ for the actual (observed) values are close to those actual values, we use mean absolute error between the actual values and the corresponding generated values. This loss is defined as the *masked reconstruction loss* ($loss_R$). To calculate this loss, we mask the input $x$ and output $\tilde{x}$ of the generator:

$$loss_R = x \odot m - \tilde{x} \odot m \quad (8)$$

We also use a *consistency loss* ($loss_c$), which is the difference between the forward $\widetilde{x^f}$ and backward $\widetilde{x^b}$ generated values:

$$loss_c = \widetilde{x^f} - \widetilde{x^b} \quad (9)$$

The discriminator $D$, also consists of one bidirectional recurrent layer of LSTM cells. Using a binary cross-entropy loss, we train $D$ to maximize the probability of correctly classifying the actual (as real) and the generated values (as fake).

$$loss_D = -\log(D(\bar{x} \odot m)) - \log(1 - D(\bar{x} \odot (1 - m))) \quad (10)$$

where $loss_D$ is the classification loss for the discriminator. $D(\bar{x} \odot m)$ is the probability of an actual value being classified as real, and $1 - D(\bar{x} \odot (1 - m))$ is the probability of a generated value being classified as fake. In a way, $D$ is trained to correctly reproduce the mask vector.

We simultaneously train $G$ to minimize $\log(1 - D(\bar{x}))$. $G$ will learn to decrease the probability that $D$ correctly identifies the fake instances:

$$loss_G = \log(1 - D(\bar{x} \odot (1 - m))) \quad (11)$$

where $loss_G$ is the classification loss for the G. Lowering $loss_G$ equals decreasing the probability that D classifies fake instances as fake.

We accumulate all the losses for $G$ and $D$ as follows:

Generator Loss = $loss_R + loss_c + loss_G$

Discriminator Loss = $loss_D$

Figure 1 shows all these losses by blue lines, while the red lines show the backpropagation of the calculated loss values.

### Imputation and Prediction Implementation

After presenting the structure of Bi-GAN, we can now explain how it can be used for both imputation and prediction tasks. The model is always trained using the imputation setting. For training the model, we align all the consecutive time-series for all samples. Shorter length time-series are padded with zeros. These padded values are treated as missing values. The model learns to estimate the values from all the samples irrespective of the length of the

---

https://github.com/mehak25/BiGAN

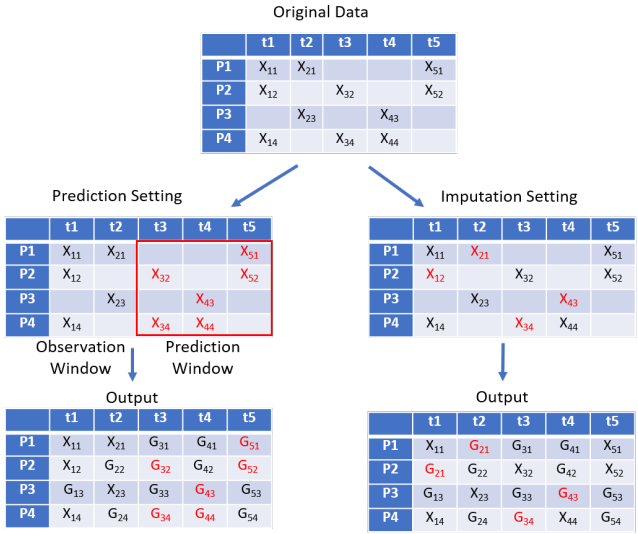

Figure 2: Imputation and Prediction settings for Bi-GAN

time-series. It estimates the missing values in-between or outside the observed time-series. During the test phase, we can use two different settings – imputation and prediction to evaluate the model's performance. Figure 2 shows the imputation and prediction settings for Bi-GAN, where the original (input) data only consists of the values of the univariate target variable that is being imputed for the samples *P1* to *P4* on time steps *t1* to *t5*. Missing values are represented by = empty cells. In the imputation setting (the right side in Figure 2), a few values are randomly deleted from the original matrix represented in red. The model fills-in all the missing and deleted values. The *masked reconstruction loss* ($loss_R$) reported as the imputation performance of the model is then calculated for the values represented in red.

In the prediction setting (for the test phase), we divide the time-series data into observation and a prediction window with the desired length. In Figure 2, we set the observation window to include *t1* and *t2* and the prediction window to include *t3* to *t5*. We delete all the observed values in the prediction window shown in red. Model fills-in all the missing and deleted values as shown in the corresponding output matrix. The *masked reconstruction loss* ($loss_R$) reported as the prediction performance of the model is calculated for the values represented in red.

## Experiments

### Datasets

**Air Quality** – This is a multivariate time-series dataset that contains 9,358 instances of hourly averaged responses from an array of 5 metal oxide chemical sensors embedded in an Air Quality Chemical Multisensor Device within an Italian city. The dataset is available on the UCI repository (De Vito et al. 2008). Data were recorded from March 2004 to February 2005 (one year). Hourly response averages are recorded along with gas concentrations references from a certified analyzer. The total number of attributes in the data is 15 and they are recorded as real values. Missing values are tagged with a "-200" value. Overall, there are 13.7% of the values that are missing. We divided the data for each month into samples with 20 consecutive observations each. We used this data to impute and predict CO(GT) values from the multivariate time-series. CO(GT) values have 18% missing rate. We have used the data for the months of July 2004, Oct 2004, and Feb 2005 as the test set; May 2004 as the validation set, and the rest of the data as the train set.

**MIMIC-III** – MIMIC-III is a large, freely-available database comprising deidentified health-related data associated with over 40,000 patients who stayed in critical care units of the Beth Israel Deaconess Medical Center between 2001 and 2002. We used the lab values data table that contains all the laboratory measurements for the patients. This is the same cohort as the one derived in other studies (Raghu et al. 2017). Overall, there are 81.9% of the values that are missing. We took 20 consecutive values for each patient. We post-padded the time-series for the patients with less than 20 consecutive time-series observations. We use this data to impute and predict WBC (white blood cell) values from the multivariate time-series. WBC values have 80% missing rate. We divided the samples in the dataset into 70:10:20 as train, validation and test sets.

**Electronic Health Record Data** – The EHR data used in this work was extracted from a large pediatric health system. A more detailed description of the dataset and our preprocessing steps for preparing the raw EHR data are presented in earlier work (Gupta et al. 2019). All the data access and processing steps were approved by an institutional review board. The EHR dataset consisted of about 44 million records with 68,029 distinct patients and 34,96,559 distinct visits. The final cohort used in this study consisted of 66,878 patients and 607 medical variables. Medical variables consisted of the condition and measurement variables. In this study, we used 0-10 years of data for the patients. We represented all the conditions as binary variables (1 if present, and 0 if not recorded for the visit). Measurement (continuous) variables were normalized. For using this EHR dataset in an RNN-based deep learning model, we converted the time-series data from irregularly-spaced time intervals to regular time-series with equal time intervals. We aggregated the time-series data over a 6-month period. By averaging values for time-series data over the 6-month periods, we have obtained 20 timestamps for each patient for the period of 0 to 10 years of age. If a patient did not have any visit over a certain 6-month period, that visit's entries were padded with zeros.

https://github.com/mehak25/BiGAN

We use this data to impute and predict patients' BMI values for the next 5, 6, 7, and 8 years. Missing ratio for the BMI values is 72% in the data.

**Experiment Setting**

Using the datasets explained above, first, we evaluate the imputation and prediction performance of Bi-GAN in comparison to other similar state-of-the-art methods. Second, we evaluate the imputation performance of Bi-GAN with various missing rates, and its prediction performance with different observation and prediction window lengths. Finally, we qualitatively analyze the properties of Bi-GAN by evaluating the effects of its components on its performance.

We conduct our experiments using 5-fold cross-validations. We report MAEs as the performance metrics with 95% confidence intervals. Following the method described above, for the training phase, missingness is applied by randomly removing 10% of data points, and prediction is performed using the first 4 consecutive time-series as and observation window and the last 16 consecutive time-series as the prediction window. The models will predict all the 16 values in the prediction window. Complete code is available at https://github.com/mehak25/BiGAN.

**Experiment Results**

**Quantitative Analysis**

We have used the above three datasets to quantitatively evaluate the performance of Bi-GAN. In Table 1 and Table 2, we report the MAE scores for Bi-GAN and four other imputation methods: BRITS-I , MICE (Azur, Stuart, Frangakis and Leaf 2011), KNN, and Mean (imputing by the mean value). We tested all methods in both imputation and prediction settings. Table 1 shows that Bi-GAN outperforms all other methods in imputation settings. Table 2 shows that Bi-GAN achieves the best prediction performance compared to the other methods in EHR and MIMIC datasets, and a similar performance to Bi-GAN on the air quality dataset. When using the model for prediction, we delete all the values in the prediction window. Since all the values on the right side of the time series are deleted, it is expected that the values generated by the bi-direction recurrent dynamics in the generator will be much less accurate in the backward direction than the forward direction. By using the combination factors, $\lambda^f$ and $\lambda^b$, our model is able to learn the best way for combining the forward and backward imputed values in both imputation and prediction setting. State-of-the-art methods such as BRITS-I use the mean of forward and backward imputed values, which can obtain a poor performance (especially in the prediction setting for datasets with a large number of variables), since the backward imputed values are much more inaccurate than the forward imputed values. Mean, KNN, and MICE also do not work well for prediction setting, because they rely on the observed values in other rows and in the same column to predict missing values, while in the prediction setting all the columns in the prediction window are empty.

Table 1: Imputation performance in MAE (95% CI)

| Algorithm | EHR | Air Quality | MIMIC |
|---|---|---|---|
| Bi-GAN | **1.23** (0.012) | **1.12** (0.24) | **7.41** (0.32) |
| BRITS-I | 2.78 (0.04) | 1.74 (0.20) | 9.33 (0.27) |
| MICE | 2.167 | 34.97 (26.84) | 4.15 (0.17) |
| KNN | 2.24 (0.02) | 15.16 (10.25) | 4.22 (0.24) |
| MEAN | 1.78 (0.02) | 54.06 (1.00) | 5.27 (0.23) |

Table 2: Prediction performance in MAE (95% CI)

| Algorithm | EHR | Air Quality | MIMIC |
|---|---|---|---|
| Bi-GAN | **1.99** (0.01) | 70.97 (3.72) | **9.67** (0.21) |
| BRITS-I | 7.65 (0.01) | **70.79** (3.75) | 10.54 (0.22) |
| MICE | 17.38 (0.01) | 70.89 (3.78) | 12.79 (0.22) |
| KNN | 17.38 (0.01) | 70.89 (3.78) | 12.79 (0.22) |
| MEAN | 17.38 (0.01) | 70.90 (3.78) | 12.79 (0.22) |

**Imputation and Prediction Performance in different settings**

We also evaluate the performance of Bi-GAN on the EHR dataset with various missing rates in the imputation setting. Figure 3(a) shows the performance of Bi-GAN along with BRITS-I, KNN, and Mean imputation methods. We vary the missing rate from 10% to 50%. As we can see Bi-GAN outperforms all other methods throughout the range of different missing rates.

Also, in the prediction setting, we evaluate the performance of the methods by varying the observation and prediction window sizes. For time-series of length 20, we use observation window sizes of 4, 6, 8, and 10, and corresponding prediction window sizes of 16, 14, 12, and 10. As we can see from Figure 3(b), as the observation window size increases, that is the prediction window becomes shorter (moving from long-term to short-term prediction) the performance of all models increases. However, Bi-GAN outperforms all other methods.

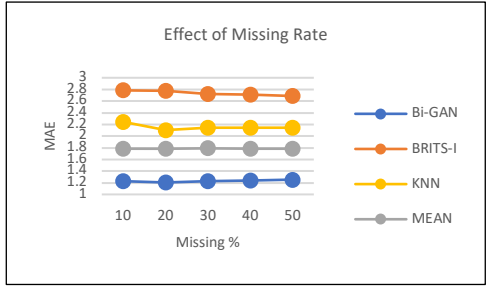

(a): MAE performance with various missing rates %

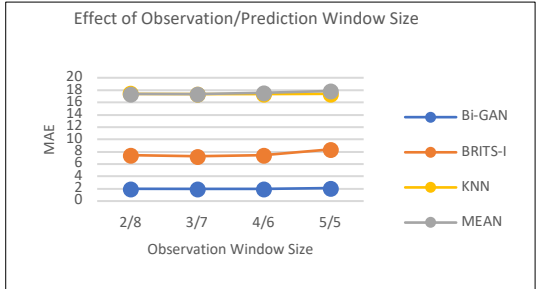

(b): MAE performance with various observation/prediction window sizes.

Figure 3: Performance comparison in different imputation and prediction settings

**Model Component Performance**

The potential components that added performance boost to Bi-GAN compared to the state-of-the-art methods include the use of GAN like architecture by adding a discriminator and adding losses $L_G$ and $L_D$, as well as the use of $\lambda^f$ and $\lambda^b$ to combine the values obtained from forward and backward directions. Our model without the discriminator and the combination factors ($\lambda^f$ and $\lambda^b$) is comparable to BRITS-I (Cao, Wang, Li, Zhou, Li and Li 2018) model without the time-series classification layer. We also compared the results of the GAN model by using Wasserstein (Arjovsky et al. 2017) loss, instead of a vanilla loss (binary cross-entropy loss). Table 3 shows that the performance is best in both imputation and prediction setting when all the components are included.

Table 3: Model Component Performance on EHR data (95% CI)

| Algorithm | Imputation MAE (95% CI) | Prediction MAE (95% CI) |
|---|---|---|
| Bi-GAN | **1.2327 (0.0123)** | **1.988 (0.0078)** |
| Discriminator with Wasserstein loss | 1.772 (0.0177) | 2.061 (0.0081) |
| w/o $\lambda^f, \lambda^b$ | 2.719 (0.0198) | 5.23682 (0.0106) |
| w/o $loss_G, loss_D$ | 1.28 (0.0174) | 2.25 (0.0082) |
| w\o $loss_G, loss_D$ and $\lambda^f, \lambda^b$ | 2.78 (0.037) | 7.65 (0.0122) |

https://github.com/mehak25/BiGAN

## Conclusion

In this study, we have developed a generative adversarial network with bi-directional recurrent units for both imputation and prediction on time-series data. We have demonstrated that our proposed model outperforms other state-of-the-art models, which are used for both imputation and prediction tasks. Our proposed model could achieve imputation and prediction MAE of 1.23 and 1.99 respectively on a real-world EHR dataset for imputing and predicting BMI values in children of 0-10 years of age, compared to the next best method yielding MAEs of 2.78 and 7.65. By approaching the task of prediction as to the imputation of future values, we were able to achieve a flexible prediction window for both short-term and long-term predictions. Our method can additionally work on time-series data of varying lengths by treating shorter time-series as samples with missing values.

https://github.com/mehak25/BiGAN